# Comparing Surface Landmine Object Detection Models on a New Drone Flyby Dataset

Navin Agrawal-Chung
*Monta Vista High School*
Cupertino, CA, United States
navin@unveilx.org

Zohran Moin
*Mountain View High School*
Mountain View, CA United States
zohran@unveilx.org

*Abstract - **Landmine detection using traditional methods is slow, dangerous and prohibitively expensive. Using deep learning-based object detection algorithms drone videos is promising but has multiple challenges due to the small, soda-can size of recently prevalent surface landmines. The literature currently lacks scientific evaluation of optimal ML models for this problem since most object detection research focuses on analysis of ground video surveillance images. In order to help train comprehensive models and drive research for surface landmine detection, we first create a custom dataset comprising drone images of POM-2 and POM-3 Russian surface landmines. Using this dataset, we train, test and compare 4 different computer vision foundation models YOLOF, DETR, Sparse-RCNN and VFNet. Generally, all 4 detectors do well with YOLOF outperforming other models with a mAP score of 0.89 while DETR, VFNET and Sparse-RCNN mAP scores are all around 0.82 for drone images taken from 10m AGL. YOLOF is also quicker to train consuming 56min of training time on a Nvidia V100 compute cluster. Finally, this research contributes landmine image, video datasets and model Jupyter notebooks at* https://github.com/UnVeilX/ *to enable future research in surface landmine detection**.*

## 1. INTRODUCTION

Landmines have caused more than 1,000,000 casualties in the last 50 years. Since surface landmine dispersal rates are 25 times faster than they be cleared, faster and safer mine detection techniques are needed. 175,000 sq-km or 25% of occupied Ukraine has been mined since the start of the Ukraine war and the UN estimates that demining will require over 100 years and more than $37 billion [1].

These landmines cause deaths and injuries, make agricultural lands unsafe, and prevent displaced individuals from returning back home. Current landmine detection strategies include the use of manual metal detectors and armored demining vehicles, which have proven to not only be dangerous, but both inefficient and costly as it takes more than 7 months and $150,000 to train a bomb specialist. Some organizations have tried using aerial approaches like satellites [2] but with limited success due to the limited resolution of satellite images and inability to detect the smaller soda-can size POM-2 and POM-3 [4] surface mines prevalent in Ukraine. This is because satellite image resolution is typically no better than 5m per pixel. Even in the cases when 25cm per pixel is possible, the size of surface landmines makes this a challenging problem for satellites.

Use of drones is a new method [5] that is only recently being explored but it has its own challenges due to lack of datasets, trained models and practical solution guidelines that allow for detection of these smaller, surface mines. Further challenges include wide-vision and arbitrary orientation for the drone camera, limited compute availability on airborne platforms and complex backgrounds. The use of drone flyby videos in computer vision machine learning models aims to assist demining efforts by eliminating the need for ground-based human labor for mine detection.

## 2. RELATED WORK

In recent months, we have seen multiple efforts at solving the landmine detection problem in Ukraine. Jasper Baur reports a landmine detection software solution developed by Safe Pro group and John Frucci at the University of Oklahoma OSU Global Consortium for Explosive Hazard Mitigation [1]. Their SpotlightAI application takes in imagery from commercial off-the-shelf-drones and uses an AI model to detect certain mine classes.

However, currently published studies have not compared the accuracy of different computer vision models, model training and inferencing, runtime and optimal model selection to quantitatively characterize which is the optimal model from point of view of inference speed and accuracy for surface landmines. Moreover, a dataset on which further work can be done is also not made available.

Other approaches like in [18-29] have attempted to use thermal cameras for low light object detection. However, evaluation of thermal approaches for landmine detection are not yet available in the literature.

## 3. DATASET

Surface landmines are relatively new but unfortunately, widely in use in recent conflicts especially due to ease of large-scale remote dispersal. As a result of their relatively new deployment, datasets for surface landmines are scarce in the public domain. Therefore, our first step was to create such a dataset. A surface landmine dataset is different from buried landmines since it is possible to see such mines using visual and thermal cameras mounted on drones. Therefore, even though the speed of dispersal of such landmines is much easier than buried mines, drone datasets can help with their detection.

Our dataset is specifically built for surface landmine object class. We have built our dataset using scale models of Russian POM-2 and POM-3 surface mines. Drone images and videos were taken using a DJI-Mini3 drone at 720p and 1080p resolution.

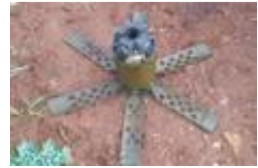

*Figure 1: POM-3 surface landmine*

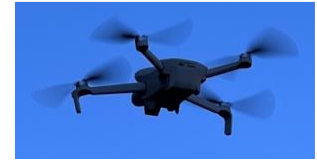

*Figure 2: DJI Mini-3 Drone*

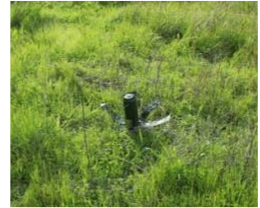

*Figure 3: POM-3 in grassy terrain*

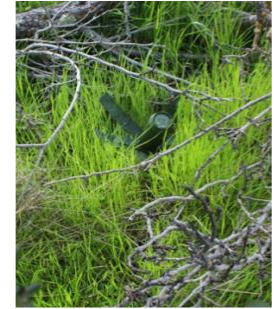

*Figure 4: POM-3 in wooded terrain*

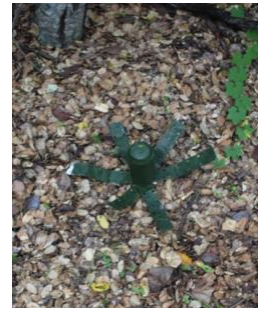

*Figure 5: POM-3 in wooded terrain*

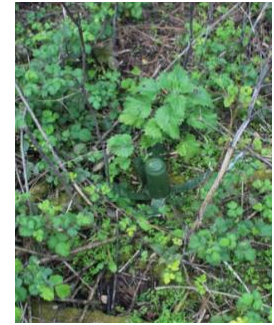

*Figure 6: POM-3 in wooded terrain*

We flew the drone at 2.5m, 5m and 10m altitudes above ground level (AGL) over different vegetation conditions, images are shown below.

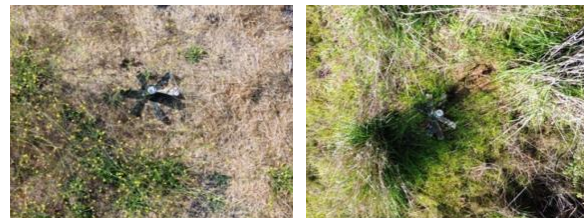

*Figure 7: View from 2.5m AGL*

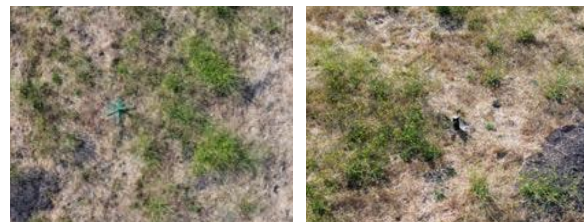

*Figure 8; View from 5m AGL*

Drone images from 10m AGL pose the greatest challenge due to the object of interest almost becoming indiscernible from the background.

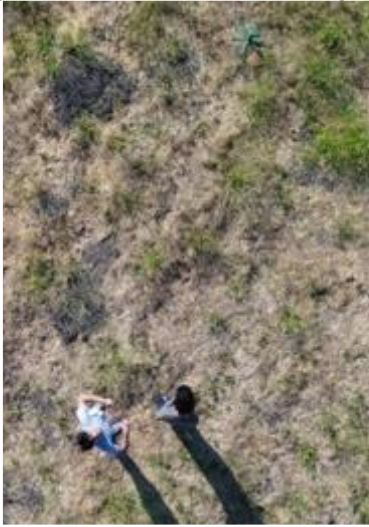

*Figure 9: View from 10m AGL*

In addition to visual images, our dataset also has a collection of 57 thermal images. These were taken with a FLIR One Edge Pro Long-Wave InfraRed (LWIR) camera with 160 x 120 pixels of resolution.

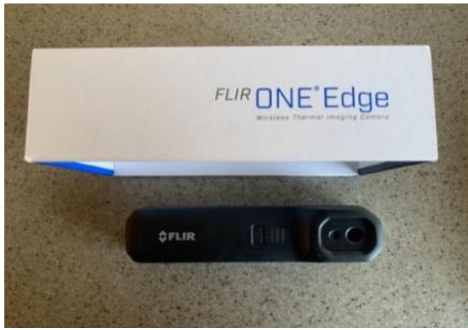

*Figure 10: FLIR One Edge PRO LWIR Camera*

Thermal cameras can allow for mine detection in low light conditions and also when the surface mine is partially obstructed by vegetation.

Our experiments showed that thermal images do not help in the presence of heavy vegetation obstruction. Longer wavelength electromagnetic waves will likely allow for detection in the presence of heavy vegetation [30]. Conducting research using different parts of the electromagnetic spectrum as a way to detect foliage obstructed landmines is proposed as a topic for future research.

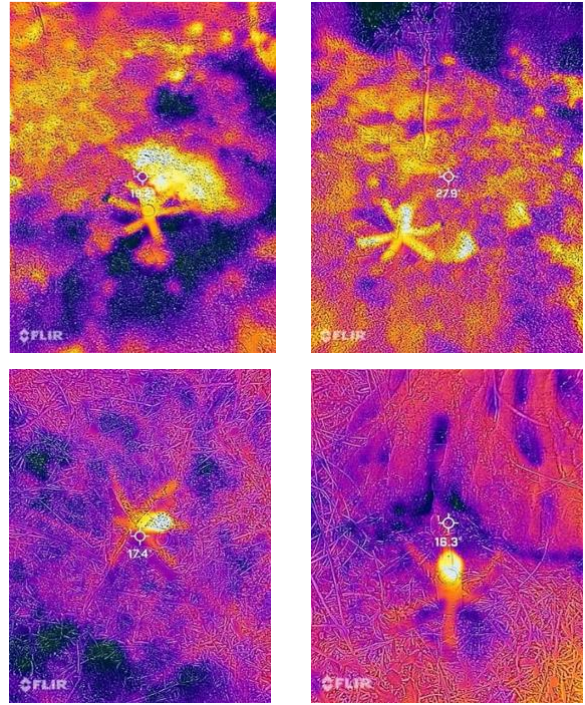

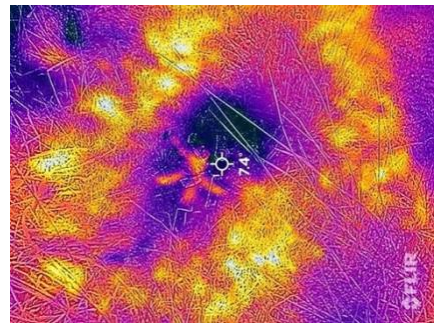

*Figure 11 POM-3 Thermal Images*

Drone speed was also varied but we found that due to high quality image stabilization built into the camera function, there was hardly any blur or image quality degradation when still images were extracted from drone videos taken anywhere from 2m/s to our drone's max speed of 15m/s.

## 4. OBJECT DETECTION MODELS AND TECHNIQUES

In our survey [7-10] we shortlisted 17 object detection models to evaluate as shown in Table 1:

| Object Detection Model | Type | Development timeline |
|---|---|---|
| YOLOV3 | One-stage | 2018 |
| YOLOv4 | One-stage | |

| YOLOv7 | One-stage | 2021 |
|---|---|---|
| YOLOv8 | One-stage | 2023 |
| **YOLOF** | **One-stage** | **2023** |
| **DETR** | **Two-stage** | **2020** |
| CPNet | One-stage | 2020 |
| NQSA | One-stage | 2022 |
| SQMFNet | One-stage | 2023 |
| **Sparse R-CNN** | **One-stage** | **2021** |
| Faster R-CNN | Two-stage | 2018 |
| RoI | Two-stage | 2019 |
| GDFNet | Two-stage | 2020 |
| MS-Faster-RCNN | Two-stage | 2021 |
| RSOD | Two-stage | 2022 |
| Fusion RCNN | Two-stage | 2023 |
| **VFNET** | **One-stage** | **2020** |

*Table 1: Model shortlist*

From this shortlist, we selected 4 models in bold above where we compare their training time on our surface landmine dataset, inference accuracy as measured by mAP scores, inference latency and other parameters like loss. We selected these models based on our assessment of whether the model has potential to be a leader in these evaluation criteria for our application as well as for model maturity. Brief descriptions of selected models are as below:

a) DETR [11] was introduced by Facebook researchers in 2020. It uses the powerful transformer architecture and is a direct-set prediction model that uses a transformer encoder-decoder to predict multiple objects at once. We wanted to include this in our model evaluation list since in many cases, surface landmines might be visually obstructed by vegetation and we wanted to

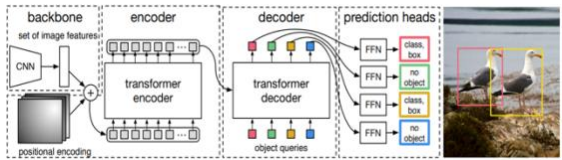

*Figure 12: DETR architecture*

see if DETR performs better detection when the mine is partially obscured by vegetation.

b) YOLO first version was first introduced in 2015 [12-13] and has undergone significant improvements till the latest version of YOLO11. For our model training, we used a variant of YOLO available on Azure cloud called YOLOF. Given that YOLO improves upon traditional sliding window CNN approaches by predicting bounding boxes and class probabilities for objects directly from the input image in a single forward pass thus making inference speed significantly faster.

c) R-CNN proposed in 2014 [14-16] solves the problem of selecting a high number of regions for proposals by limiting the number of region proposals. Given we are only interested in limited number of object classes, we wanted to include this algorithm in our evaluation set. Specifically, we considered Faster R-CNN that uses Region Proposal Networks and runs CNN once on the entire image instead of multiple times on the region of interest and thus improves on both accuracy and inference speed improvements over the original R-CNN architecture. We also considered Sparse R-CNN that uses less compute time as it generates a sparse set of learned weights, has easier training convergence due to fewer hyperparameter settings and most importantly, handles crowded scenes, for example when surface mines are obscured by lots of vegetation. However, a key limitation of R-CNN, Fast R-CNN, and Faster R-CNN is that they are relatively computationally expensive compared to other object detection methods.

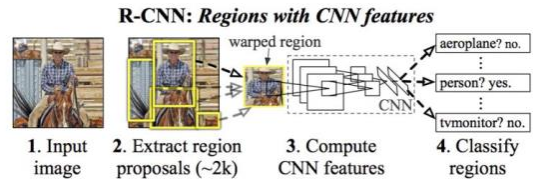

*Figure 13: R-CNN Architecture*

They also require a separate classifier for each object class, which can be computationally expensive if the number of classes is large, which fortunately in our case is not the case.

d) VFNet or VarifocalNet [17] accurately ranks a large number of candidates and is a dense object detector. It consists of CNN backbone layers, a feature pyramid and varifocal heads.

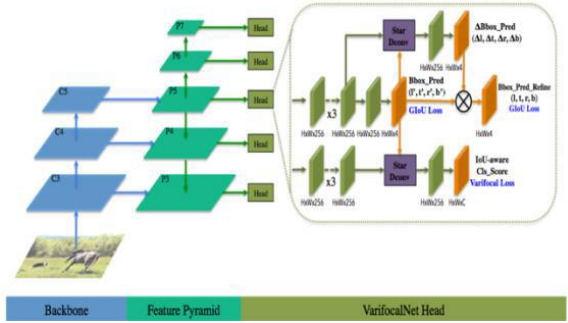
*Figure 14: VFNet Architecture*

| GPUs | 8 |
|---|---|
| Proccessor | Nvidia V100 |
| Max Network Interfaces | 8 |
| RDMA Enabled | yes |
| Accelerated Net | yes |
| OS Disk Size | 1023 GiB |
| Res Disk Size | 2900 GiB |
| Max Disks | 32 |

*Table 2: Training infrastructure*

## 5. EXPERIMENTS AND RESULTS

We trained the 4 candidate models on Azure cloud compute. 320 images were used to train the model on each model type and test data set consisted of 70 images.

Jupyter notebooks and dataset are available at https://github.com/UnVeilX/ slm-od-ml-comparison/.

Annotation for our training dataset was done manually in the Azure environment.

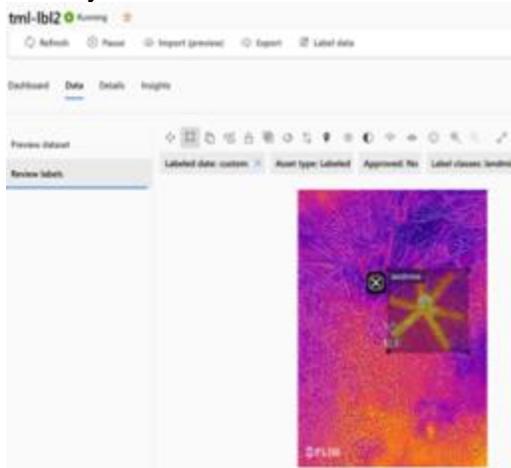
*Figure 15: Image annotation in Azure*

Training infrastructure used is described in Table 2.

| Azure Name | Standard_ND40rs_v2 |
|---|---|
| vCPUs | 40 |
| CPU Architecture | x64 |
| Memory (GiB) | 672 |
| Hyper-V Generations | V2 |
| ACUs | 0 |

Original images were 1080p resolution. They were resized to 800 * 450 resolution for training convergence.

We see in Table 3 that last loss for YOLO at 0.245 is better than the other models and considering all parameters, YOLO training metrics are overall better than the other models.

| Model | Trained model size | Training duration | Loss (last) | Training Epochs |
|---|---|---|---|---|
| YOLOF | 162MB | 56m 30s | 0.245 | 15 |
| DETR | 157MB | 1h 11m 02s | 4.2074 | 30 |
| Sparse-RCNN | 405MB | 1h 02m 28s | 4.2212 | 15 |
| VFNET | 129MB | 1h 0m 28s | 1.0257 | 30 |

*Table 3: Trained model metrics*

Figures 16-19 show comparison of various training metrics across the 4 models.

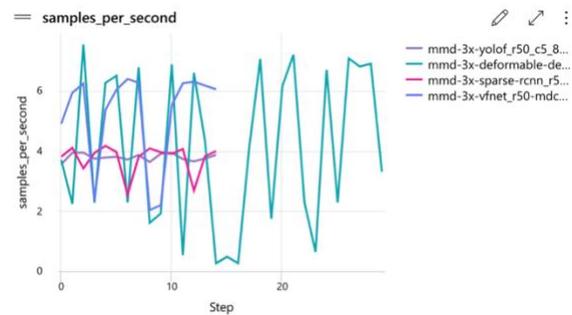
*Figure16: Comparison of training rate*

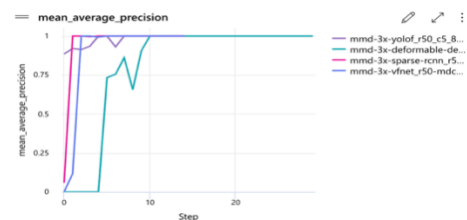
*Figure 17: Steps taken to reach final mAP*

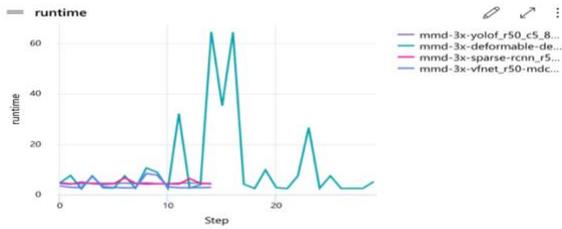

*Figure 18: Training runtime over steps*

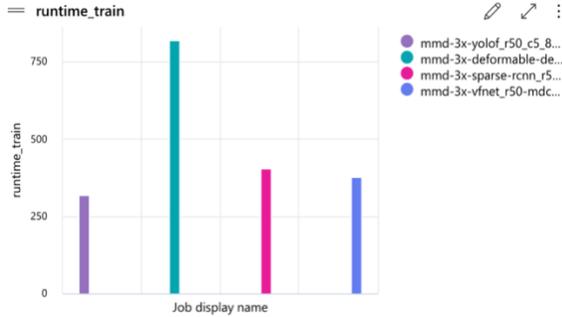

*Figure 19: Comparison of training runtime*

Inference infrastructure used is shown in Table 4.

| Name | Standard_D16as_v4 |
|---|---|
| vCPUs | 16 |
| CPU Architecture | x64 |
| Memory (GiB) | 64 |
| Hyper-V Generations | V1,V2 |
| ACUs | 0 |
| Performance Score | 292074 |
| GPUs | 0 |
| Proccessor | AMD EPYC 7452 32-Core Processor |
| NUMA Nodes | 2 |
| Max Network Interfaces | 8 |
| RDMA Enabled | no |
| Accelerated Net | yes |
| OS Disk Size | 1023 GiB |
| Res Disk Size | 128 GiB |
| Max Disks | 32 |
| Combined Write | 250 MiB/Sec |
| Combined Read | 250 MiB/Sec |

*Table 4: Inference infrastructure*

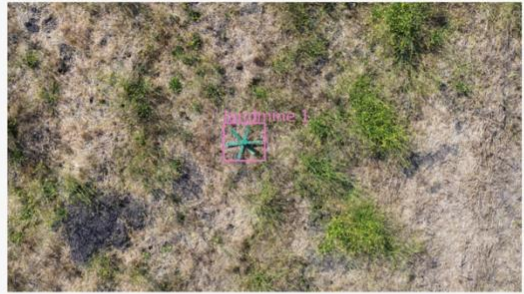

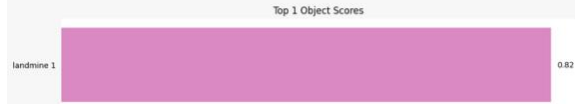

*Figure 20: DETR inference result 5m AGL, mAP=0.82*

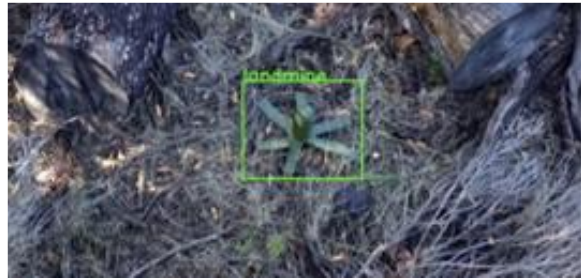

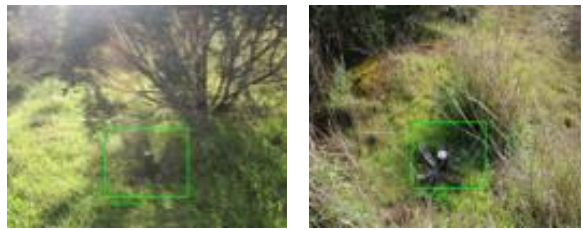

*Figure 21: Inference results on 2.5m AGL*

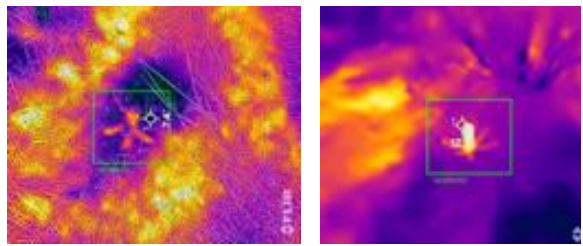

*Figure 22: Inference on thermal images*

We see in Table 5-6 that all models perform well for 2.5m and 5m AGL. Table 7 shows that for 10m AGL, when inference is performed using original sized images (without reduced resolution) since the landmines are too small to be detected from the resized lower resolution images used for training, YOLOF outperforms other models.

| Model | Inference mAP | Inference time |
|---|---|---|
| YOLOF | 0.97 | 0.6s |
| DETR | 0.98 | 1.4s |
| Sparse-RCNN | 0.97 | 0.6s |
| VFNET | 0.97 | 0.6s |

*Table 5: Inference results on 2.5m AGL drone images*

| Model | Inference mAP | Inference time |
|---|---|---|
| YOLOF | 0.97 | 0.6s |
| DETR | 0.98 | 1.4s |
| Sparse-RCNN | 0.96 | 0.6s |
| VFNET | 0.97 | 0.6s |

*Table 6: Inference results on 5m AGL drone images*

| Model | Inference mAP | Inference time |
|---|---|---|
| YOLOF | 0.89 | 0.8s |
| DETR | 0.82 | 3s |
| Sparse-RCNN | 0.83 | 1.2s |
| VFNET | 0.81 | 1.3s |

*Table 7: Inference results on 10m AGL drone images*

## 6. CONCLUSIONS

In this paper, we have trained and evaluated 4 object detection models namely YOLOF, DETR, Sparse-RCNN and VFNET for surface landmine detection on drone flyby videos. We created our own dataset for training and testing since no dataset was available in the literature. We conclude that all 4 models are generally good detectors when drones are flown between 2.5m and 5m AGL and in the presence of light vegetation. When drones are flown higher at 10m AGL, YOLOF outperforms other models. Drone speed had no influence on detection accuracy since even at max drone speed of 10m/s for which we captured videos and images, there was no image blurring. We also found that YOLOF hyperparameters were easier to setup for training convergence, model size is slightly smaller and training time is slightly faster. Regarding thermal image inferencing, we could only inference on 2.5m AGL images since thermal camera we used had limited resolution. Further research is needed for methods to improve mine detection in the presence of heavy vegetation as our results using thermal LWIR images and videos were inconclusive.

## ACKNOWLEDGEMENT

This work would not have been possible without the generous support of Microsoft for Startups Founders Hub for funding the cloud compute to train our models, access to base foundation models and technical guidance.

## REFERENCES

1. Baur, Jasper, "Ukraine is littered with landmines, Drones can help", IEEE Spectrum, 23 April, 2024.

2. Jagula, Dexter, "Satellite Imagery for Everyone", IEEE Spectrum, Feb 2022; https://spectrum.ieee.org/commercial-satellite-imagery

3. Hui, X., Bian, J., Yu, Y., Zhao, X. & Tan, M., "A Novel Autonomous Navigation Approach for UAV Power Line Inspection", IEEE International Conference on Robotics and Biomimetics, December 2017; https://ieeexplore.ieee.org/document/8324488

4. "POM-2 Mine", Wikipedia, Last edited 19 December, 2023; https://en.wikipedia.org/wiki/POM-2_mine

5. Kovacs, Z. & Ember, I., "Landmine Detection with Drones", Sciendo, Vol 27, Issue 1, March 2022. Pp 84-92; https://sciendo.com/article/10.2478/raft-2022-0012?content-tab=abstract

6. Ong, Amandas, "High Tech Tools Join the Hunt for Forgotten Mines", Bloomberg News, April 22, 2021; https://www.bloomberg.com/news/articles/2021-04-23/the-delicate-art-and-science-of-mine-removal

7. Liu, Yu Tong, "The Ultimate Guide to Video Object Detection", Towards Data Science, May 13, 2020: https://towardsdatascience.com/ug-vod-the-ultimate-guide-to-video-object-detection-816a76073aef

8. Ross Girshick, Jeff Donahue, Trevor Darrell, Jitendra Malik, "Rich feature hierarchies for accurate object detection and semantic segmentation", Oct 2014; https://arxiv.org/abs/1311.2524

9. Sun, C., Zhan, W.,She, J. & Zhang, Y., "Object Detection from the Video Taken by

Drone via Convolutional Neural Networks", Hindawi Problems in Engineering, Volume 2020, Oct 2020; https://www.hindawi.com/journals/mpe/2020/4013647/

10. The Tenyks Blogger, "Computer Vision Pipeline v2.0", Medium blog, Jan 11, 2024; https://medium.com/@tenyks_blogger/computer-vision-is-already-evolving-3cd0e63e805b

11. Nicolas Carion, Francisco Massa, Gabriel Synnaeve, Nicolas Usunier, Alexander Kirillov, Sergey Zagoruyko, "End-to-End Object Detection with Transformers", May 2020; https://arxiv.org/abs/2005.12872

12. Joseph Redmon, Santosh Divvala, Ross Girshik, Ali Farhadi, "You Only Look Once: Unified, Real-time Object Detection", June 2015; https://arxiv.org/abs/1506.02640

13. Wang, C.-Y., Bochkovskiy, A. & Liao, H.-Y. M. YOLOv7: Trainable bag-of-freebies sets new state-of-the-art for real-time object detectors. http:// arxiv.org/ abs/ 2207. 02696 (2022).

14. Gandi, R. R-CNN, Fast R-CNN, Faster R-CNN, YOLO: Object Detection Algorithms | by Rohith Gandhi | Towards Data Science. Towards Data Science https:// towardsdatascience.com/r-cnn-fast-r-cnn-faster-r-cnn-yolo-object-detection-algorithms-36d5357136 5e? gi= b2d45 005e9 a2.

15. Peize Sun, Rufeng Zhang, Yi Jiang, Tao Kong, Chenfeng Xu, Wei Zhan, Masayoshi Tomizuka, Lei LI, Zehuan Yuan, Changhu Wang, Ping Luo, "Sparse R-CNN: End-to-End Object Detection with Learnable Proposals", CVPR 2021; https://paperswithcode.com/paper/sparse-r-cnn-end-to-end-object-detection-with

16. "Object Detection using Faster-RCNN", Microsoft Learn, October 13, 2022, https://learn.microsoft.com/en-us/cognitive-toolkit/object-detection-using-faster-r-cnn

17. Haoyang Zhang, Ying Wang, Feras Dayoub, Niko Sunderhauf, "VarifocalNet: An IoU-aware Dense Object Detector", Aug 2020; https://arxiv.org/abs/2008.13367v2

18. Gallagher, J. & Oughton, E., "Assessing thermal imagery integration into object detection methods on ground-based and air-

based collection platforms", arxiv, December 2022; https://arxiv.org/abs/2212.12616

19. Bhattarai, M. & Martinez-Ramon, M., "A Deep Learning Framework for Detection of Targets in Thermal Images to Improve Firefighting", arxiv, April 2020; https://arxiv.org/pdf/1910.03617.pdf

20. St-Laurent, X., "Combination of colour and thermal sensors for enhanced object detection," 10th International Conference on Information Fusion, July 2007, pg 1–8

21. Fei, S., Hassan, M.A., Xiao, Y., Su, X., Chen, Z., Cheng, Q. Duan, F., Chen, R. & Ma, Y., "UAV-based multi-sensor data fusion and machine learning algorithm for yield prediction in wheat", Precis. Agric, 2022; https://link.springer.com/article/10.1007/s11119-022-09938-8

22. Jiang, C., Ren, H., Ye, X., Zhu, J., Zeng, H., Nan, Y., Sun, M., Ren, X. & Huo, H., "Object detection from UAV thermal infrared images and videos using YOLO models" Int. J. Appl. Earth Obs. Geoinf. 2022, 112, 102912; https://www.sciencedirect.com/science/article/pii/S1569843222001145

23. De Oliveira, D.C. & Wehrmeister, M.A., "Using deep learning and low-cost RGB and thermal cameras to detect pedestrians in aerial images captured by multirotor UAV" Sensors 2018, 2244; https://www.mdpi.com/1424-8220/18/7/2244

24. Krišto, M., Ivasic-Kos, M. & Pobar, M., "Thermal object detection in difficult weather conditions using YOLO" IEEE Access 8,125459–125476 (2020).

25. Luo, Y., Remillard, J. & Hoetzer, D. "Pedestrian detection in near-infrared night vision system" 2010 IEEE Intelligent Vehicles Symposium 51–58. https:// doi. org/ 10. 1109/ IVS. 2010. 55480 89 (2010).

26. Setjo, C. H., Achmad, B., & Faridah., "Thermal image human detection using Haar-cascade classifier", 2017 7th InternationalAnnual Engineering Seminar (InAES) 1–6. https://ieeexplore.ieee.org/document/8068554

27. Dai, X., "Object detection in thermal infrared images for autonomous driving", Appl Intell 51, 1244–1261: https://link.springer.com/article/10.1007/s10489-020-01882-2

28. Batchuluun, G. et al, "Deep learning-based thermal image reconstruction and object detection", IEEE Access 9, 5951–5971 (2021).

29. Blythman, R. et al. Synthetic thermal image generation for human-machine interaction in vehicles. In 2020 Twelfth International Conference on Quality of Multimedia Experience (QoMEX) 1–6. https:// doi.org/ 10. 1109/ QoMEX 48832. 2020. 91231 35 (2020).

30. Armed Services Technical Information Agency, "Thermal Emissivities of Some Metallic and Non-Metallic Surfaces over the Range of Temperature 70C to 250C", Unclassified reprint, AD295648, Aug 2008; https://apps.dtic.mil/sti/tr/pdf/AD0295648.pdf